\documentclass[11pt,a4paper]{article}
\usepackage[hyperref]{acl2019}
\usepackage{times}
\usepackage{latexsym}

\usepackage{url}
\usepackage{booktabs} 
\usepackage{times}
\usepackage{latexsym}
\usepackage{booktabs}
\usepackage{amssymb}
\usepackage{dsfont}
\usepackage{caption}
\usepackage{multirow}
\usepackage{subfig}
\usepackage{amsmath}
\usepackage{todonotes}
\usepackage{multirow}
\usepackage{graphicx,wrapfig,lipsum}
\usepackage{subfig}
\usepackage{csquotes}
\usepackage{hyperref}
\usepackage{url}

\usepackage{arydshln}
\usepackage[markup=underlined]{changes}

\usepackage{xcolor,colortbl}
\usepackage{array}
\newcolumntype{L}[1]{>{\raggedright\let\newline\\\arraybackslash\hspace{0pt}}m{#1}}
\newcolumntype{C}[1]{>{\centering\let\newline\\\arraybackslash\hspace{0pt}}m{#1}}
\newcolumntype{R}[1]{>{\raggedleft\let\newline\\\arraybackslash\hspace{0pt}}m{#1}}

\aclfinalcopy 

\title{Regularization Advantages of Multilingual Neural Language Models for Low Resource Domains} 

\newcommand*{\affaddr}[1]{#1} 
\newcommand*{\email}[1]{\texttt{#1}}

\author{%
Navid Rekabsaz, Nikolaos Pappas, James Henderson, Banriskhem K. Khonglah, Srikanth Madikeri\\
\affaddr{Idiap Research Institue}\\
\email{\{first\_name.family\_name\}@idiap.ch}\\
}

\date{}

\begin{document}
\maketitle

\begin{abstract}

Neural language modeling (LM) has led to significant improvements in several applications, including Automatic Speech Recognition.  However, they typically require large amounts of training data, which is not available for many domains and languages.  In this study, we propose a multilingual neural language model architecture, trained jointly on the domain-specific data of several low-resource languages. The proposed multilingual LM consists of language specific word embeddings in the encoder and decoder, and one language specific LSTM layer, plus two LSTM layers with shared parameters across the languages. This multilingual LM model facilitates transfer learning across the languages, acting as an extra regularizer in very low-resource scenarios. We integrate our proposed multilingual approach with a state-of-the-art highly-regularized neural LM, and evaluate on the conversational data domain for four languages over a range of training data sizes. Compared to monolingual LMs, the results show significant improvements of our proposed multilingual LM when the amount of available training data is limited, indicating the advantages of cross-lingual parameter sharing in very low-resource language modeling.

\end{abstract}

\section{Introduction}
\label{introduction}
Language modeling (LM) is a fundamental task in natural language processing which has been an essential component of several language and speech applications, most notably in machine translation~\cite{Koehn2003} and speech recognition (ASR)~\cite{deoras2011variational}. More recently, neural language models have been shown useful for transferring knowledge from large corpora to downstream tasks, including text classification, question answering and natural language inference \cite{peters18,howard18,openai18}.

However, training an effective neural LM typically requires large amount of written text in the required language and domain, which may not be readily available for many rare domains and languages. Even when there is a pre-trained language model trained on out-of-domain data available, its fine-tuning on very small validation sets is prone to over-fitting. This issue is  especially challenging for domain-specific tasks such as conversational text of low resource languages \cite{ragni2016multi}. 

A common approach to avoid over-fitting when dealing 
with limited amount of data is regularization. Recently, \citet{merity2018regularizing} demonstrated the effectiveness of regularization methods for a multi-layer LSTM language model, named AverageSGD Weight-Dropped LSTM (AWD-LSTM). Simultaneously, \citet{melis2018on} explored the effect of extensive parameter tuning in a multi-layer LSTM model, showing the competitive performance of LSTM to other proposed network architectures for language modeling. 

Beside regularization, parameter sharing between models in various domain/languages facilitates knowledge transfer across the models, and can be especially helpful in very low-resource scenarios. Multilingual training of neural networks has grown in the last few years for various language processing tasks such as machine translation~\cite{dong15,firat16,johnson2017google} and document classification~\cite{ferreira16,pappas2017multilingual}. Parameter sharing has also been studied across different tasks, such as document summarization~\cite{zhou2018neural}, reading comprehension~\cite{nishida2018retrieve}, and question answering~\cite{sachan2018self}.

\begin{figure}[t]
\centering
\includegraphics[width=0.48\textwidth]{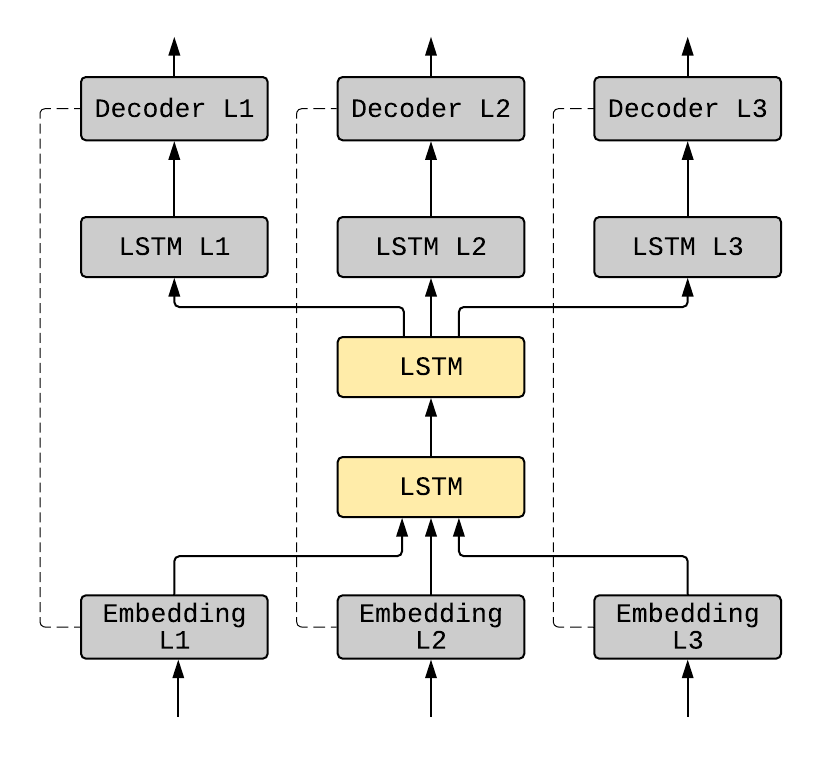}
\vspace{-10mm}
\caption{The proposed multilingual architecture.}
\vspace{-5mm}
\label{fig:mlm}
\end{figure}

In this study, we propose a 
new approach for multilingual neural language modeling along the direction of~\citet{ragni2016multi}. Our proposed multilingual architecture consists of a stacked LSTM model with three layers, where the first two layers are shared across multiple languages, and the last layer is language-specific (Figure~\ref{fig:mlm}). The first two LSTM layers capture the common patterns across languages, while the last layer learns language specific subtleties. In contrast to~\citet{ragni2016multi}, our proposed model does not need per language fine-tuning because its language-specific and language-agnostic features are simultaneously trained. Similar to previous work on multilingual training~\cite{pappas2017multilingual,firat16}, every language in our model has separate input and output layers and hence a separate loss function. We integrate the regularization methods proposed by \citet{merity2018regularizing} in our multilingual LM, resulting in a highly regularized multilingual LM, where each language benefits from the shared parameters, and, hence, the training data of other languages. 

We evaluate the proposed models in terms of perplexity on conversational data of four low-resource languages, namely Creole, Tagalog, Turkish, and Swahili, against state-of-the-art monolingual AWD-LSTM model. We study the effect of multilingual training on very low resource settings, by limiting the number of words in the training set of each language, and evaluate the performance of models over a range of available training data. Our main contributions are:
\begin{itemize}
    \item We propose a highly regularized multilingual language model for low-resource domains. 
    \item We demonstrate its superiority and stability against strong monolingual baselines when the amount of training data is very limited. 
\end{itemize}
The benefit of our regularized multilingual model is most pronounced on the Swahili language, the corpus of which is generally much smaller than the rest of the languages ($\sim$230K words). In this case, the multilingual training outperforms monolingual even on the full resource setting.




\section{Highly Regularized Multilingual Language Modeling}
\label{mlm}
\vspace{-3mm}


The proposed architecture is illustrated in Figure~\ref{fig:mlm}, for three languages \textsf{\small L1,L2,L3}. Firstly, a language-specific word embedding maps the input of the given language to its embedding vectors. Secondly, two layers of LSTMs with shared parameters capture the common patterns across the languages, followed by a language-specific LSTM for modeling language-specific characteristics. Thirdly, a language-specific decoder applies a linear transformation followed by the softmax function, and outputs the predicted probability distribution $\hat{y}^{(l)}_{t}$ at timestep $t$ over the vocabulary of the given language $l$. The weights of the input embedding and decoder for each language are tied. 

The proposed model is selected based on its superior performance in our preliminary evaluation results among other possible parameter sharing architectures with three LSTM layers, namely sharing all, only first/last, or last two layers.

\subsection{Multilingual Training}

For training, we use a training objective similar to \citet{firat16} and \citet{pappas2017multilingual}; we use a joint multilingual objective that facilitates the sharing of a subset of parameters for each language $\theta_1, \ldots, \theta_M$ of our stacked LSTMs:
\vspace{-5mm}
\begin{equation}
\vspace{-3mm}
  \label{multilingual_objective}
\begin{split}
  \mathcal{L}(\theta_1, \ldots, \theta_{M} )
 = - \frac{1}{Z}  \sum^{N_e}_{t} \sum^{M}_{l} \mathcal{H}(y^{(l)}_{t}, \hat{y}^{(l)}_{t})  
\end{split}
\end{equation}
\noindent where $M$ is the number of languages, $Z = M \times N_e$,  $N_e$ is the epoch size, and $\mathcal{H}$ is the cross-entropy loss between the ground-truth words and the predicted ones. Note that the sentence order in each language is preserved above 
and that the overall loss is back-propagated through the network, updating both language-specific and language-independent parameters. The sentences are processed in a cyclic fashion for the languages which have lesser number of sentences; once the last sentence of the text corpus is processed for that language, the next sentence that is processed is the beginning one. The joint objective $\mathcal{L}$ is minimized with respect to the parameters $\theta_1, \ldots, \theta_M$. 

\subsection{Regularization Techniques}
Pursuing the work of \citet{merity2018regularizing} on regularizing neural language models, we apply Weight-Dropped LSTM, Variational Dropout, Embedding Dropout, and Variable Length Backpropagation Sequences. \citet{merity2018regularizing} also use Activation Regularization (AR) and Temporal Activation Regularization (TAR), two weight regularization terms added to the loss function. In our multilingual LM architecture, we add these terms to the loss function of each language while their values are divided by $M$.

\section{Experiments}
\label{experiments}
\begin{table}[t]
\begin{center}
\caption{Number of words in the splits obtained from the Babel collection~\cite{gales2014speech}.}
\vspace{-0.2cm}
\begin{tabular}{l | c c c c}
 & Creole & Swahili & Tagalog & Turkish \\\hline
Train & 417539 & 237677 & 526528 & 494715 \\
Valid.  & 87418 & 6584 & 5158 & 67090 \\
Test & 84358 & 6584 & 5158 & 72106 \\
\end{tabular}
\label{tbl:collection} 
\end{center}
\vspace{-0.5cm}
\end{table}

\begin{figure*}[t]
  \centering
\subfloat[Creole]{\includegraphics[width=0.4
\textwidth]{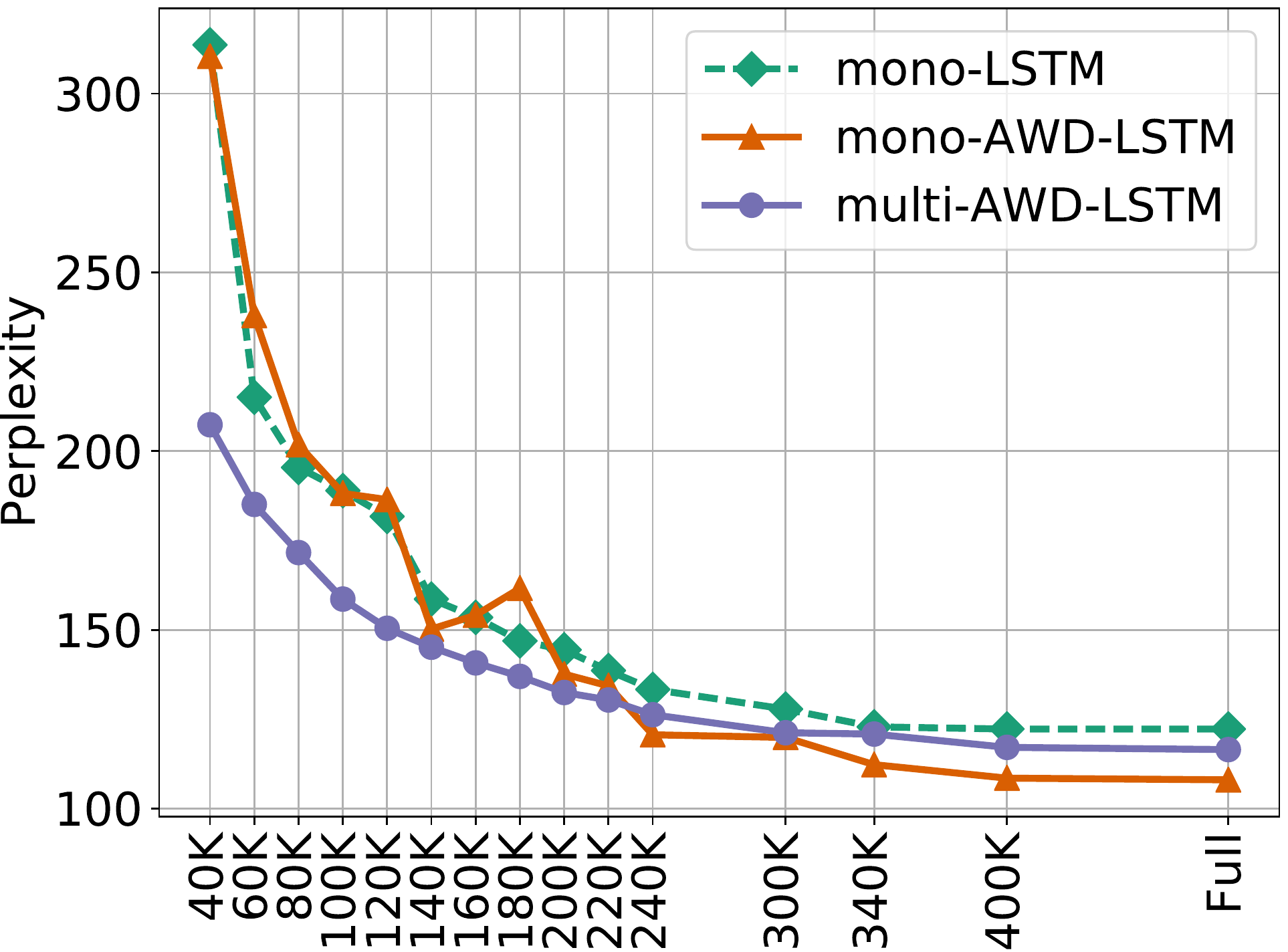}\label{fig:lm_creole}}
  \quad
\centering
 \subfloat[Swahili]{\includegraphics[width=0.4\textwidth]{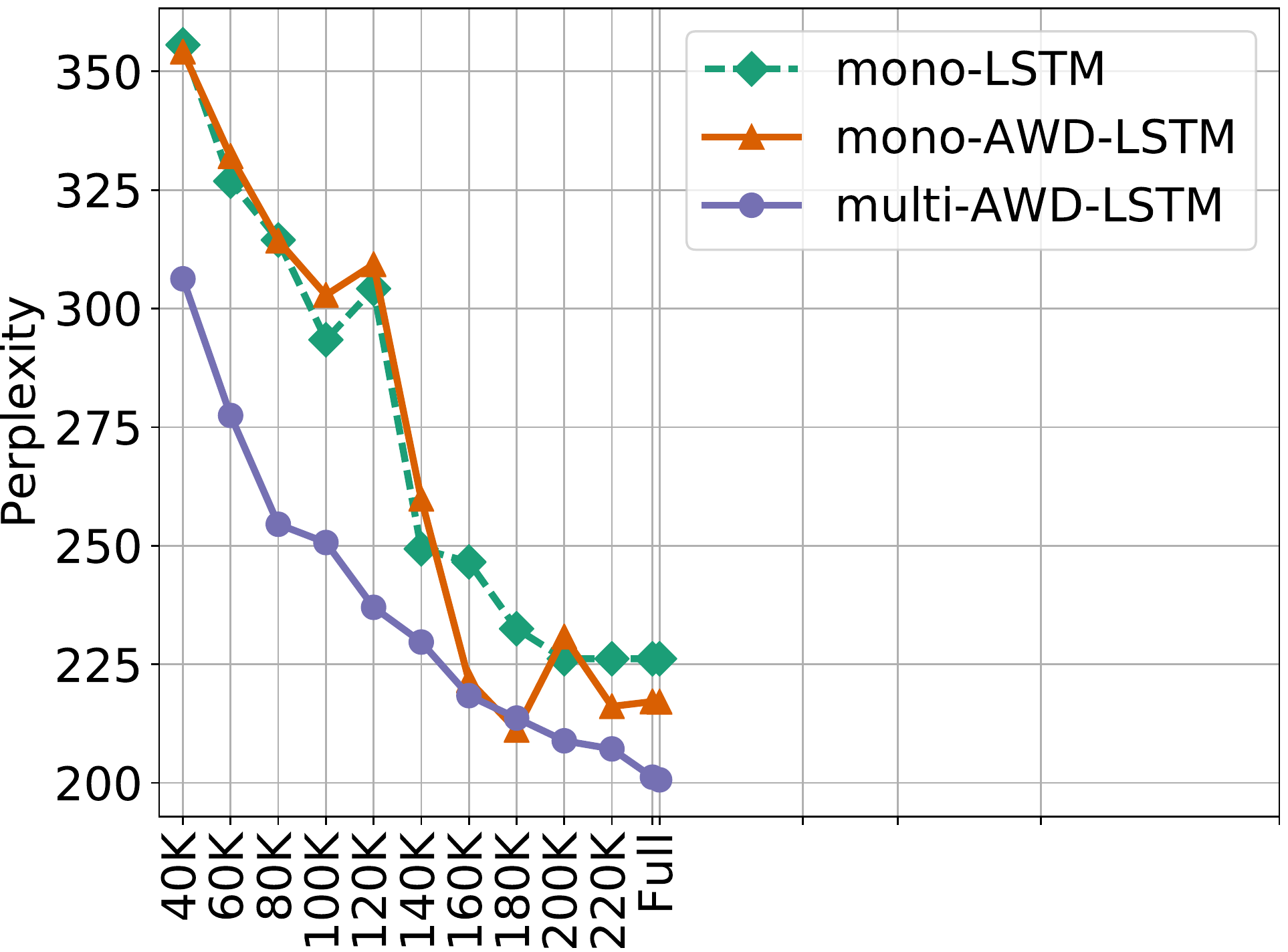}\label{fig:lm_swahili}}

\centering
\subfloat[Tagalog]{\includegraphics[width=0.4\textwidth]{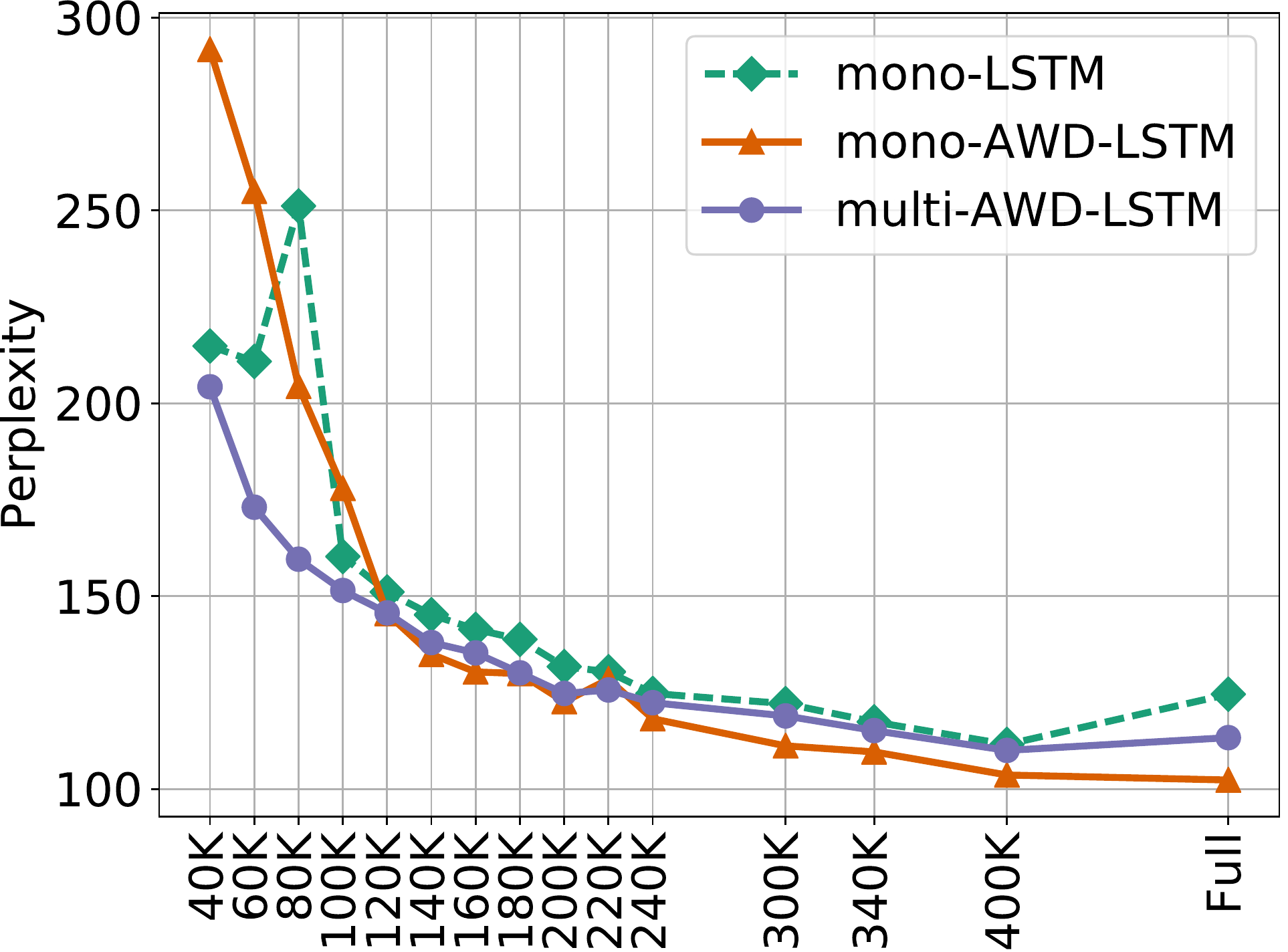}\label{fig:lm_tagalog}}
  \quad
\centering
\subfloat[Turkish]{\includegraphics[width=0.4\textwidth]{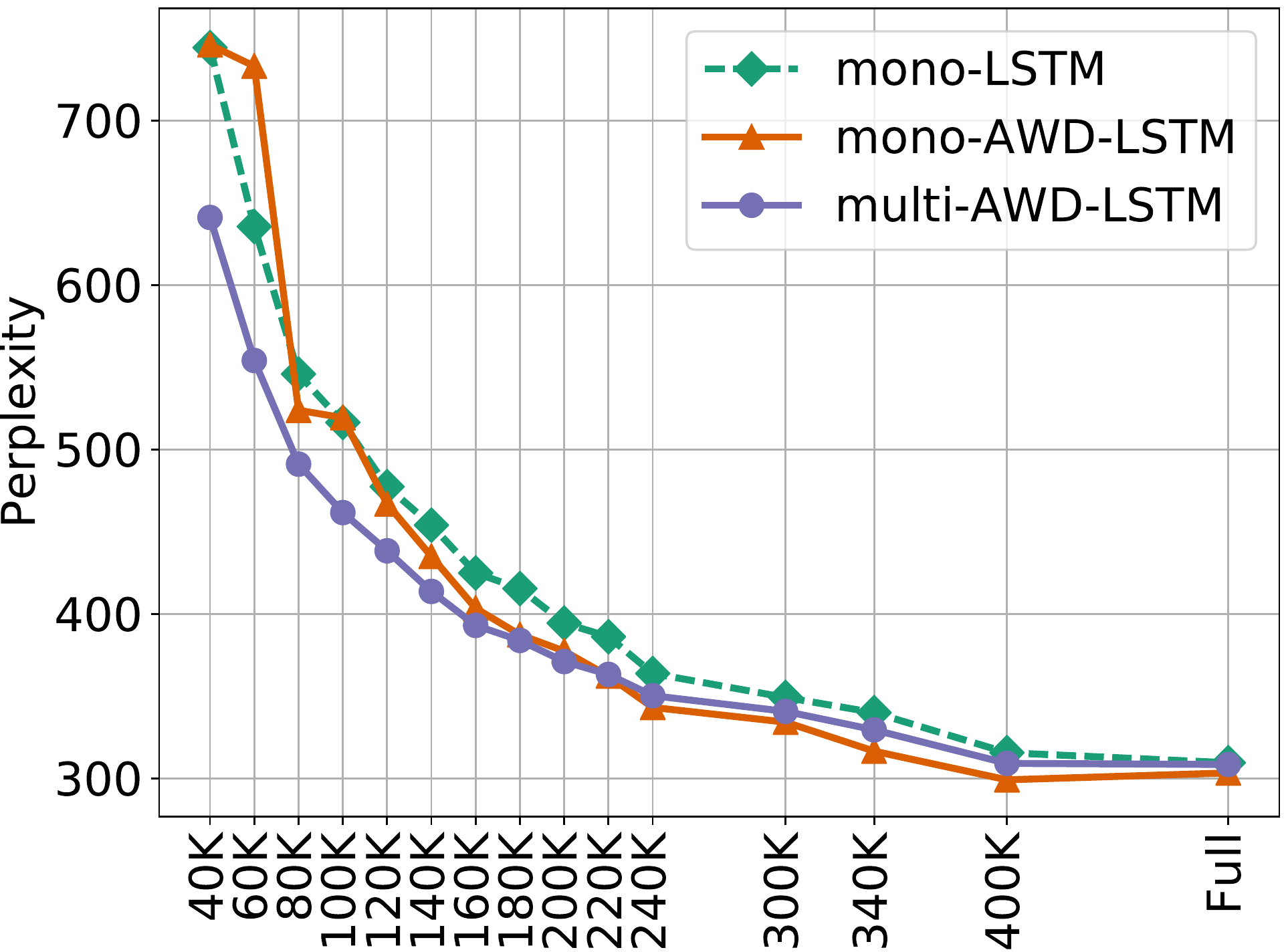}\label{fig:lm_turkish}}

 \caption{Perplexity of the three LM models on the test data of four low-resource languages. The X axis shows the size of the training data (number of words), used to learn the models. }
 \label{fig:result}
 \vspace{-5mm}
\end{figure*}

\subsection{Data and Settings}
For evaluation we use the conversational data of four low-resource languages, namely Creole, Swahili, Tagalog, and Turkish, taken from language packs released within IARPA Babel program~\cite{gales2014speech}. Every language pack contains a training and a development set, containing text of audio transcripts. We use the development set for testing and split the given training set into training and validation sets, where the size of the validation set is the same or close to the testing set. The statistics of our training/validation/test sets are reported in Table~\ref{tbl:collection}. We apply punctuation removal and set the texts to lower case. Similar to~\citet{ragni2016multi} for each language, we replace 25\% of the vocabulary words with the lowest frequencies with ${<}unk{>}$.
 
To measure the effect of data scarcity, in addition to training the models on the full texts, we also train the models on limited parts of training texts. In these scenarios, for every language only a specific number of words (based on a threshold) are used for training, selected from the beginning of the training text of that language. We train the LM models over a range of such threshold values from 20K to 400K as well as on the full training text for the languages. It should be noted that when the text of a language is restricted, the multilingual LM still have access to the full training texts of other languages.

\subsection{Model Configuration}
We set the hyper-parameters as suggested by~\citet{merity2018regularizing} for the Penn Tree Bank language modeling as follows: embedding size of 512, LSTM hidden layer size of 1150, initial learning rate to 30, batch size to 20, maximum number of epochs to 200, and sequence length of 70. The dropout rates for input, output, variational, embedding, and weight dropouts are set to 0.65, 0.4, 0.3, 0.1, 0.5, respectively. The alpha and beta values of the AR and TAR methods are set to 2 and 1. 
Lastly, we tie the weights between input embeddings and softmax weights for all the models (per language), and use  Stochastic Gradient Descent \cite{bottou10} for optimization. Our code is made available with the submission and will be made publicly available upon publication. 

The word embeddings of all the models are initialized randomly and updated during training. We also examined initializing with pre-trained cross-lingual word embeddings using the vectors provided by \citet{lample2018word} as well as creating new ones based on the unsupervised method proposed by~\citet{ammar2016massively}. In both cases, we observe the same LM performance to the models with randomly initialized embeddings, since the embeddings lose their cross-lingual alignment properties as they are being updated.

\subsection{Baselines}
We compare our multilingual model with two monolingual LSTM models: 
\begin{itemize}
    \item \texttt{mono-LSTM}: an out-of-the-box monolingual three-layer LSTM with regularization only through dropouts on the input of output layers of LSTM units.
    \item \texttt{mono-AWD-LSTM}: the same model as above but with additional regularization methods, namely Weight-Dropped LSTM, Variational Dropout, Embedding Dropout, Variable Length Backpropagation Sequences, AR, and TAR, as in~\citet{merity2018regularizing}. 
\end{itemize} 
We note as \texttt{multi-AWD-LSTM} our multilingual model with the same  regularization as \texttt{mono-AWD-LSTM}. 

\section{Results and Discussion}
\label{results}
\begin{table}[h!]
\scriptsize
\begin{center}
\caption{Perplexity of the three LMs. 
}
\vspace{-2mm}
\begin{tabular}{l | l | c c c c c }
\multirow{2}{*}{} & \multirow{2}{*}{Model} & \multicolumn{4}{c}{Number of words in training data}\\\cline{3-7}
 & & 40K & 100K & 200K & 300K & FULL \\\hline
\multirow{3}{*}{Cr} & LSTM & 313.64 & 188.96 &  144.42 & 127.84 & 122.27\\
& AWD-LSTM & 310.41 & 188.18 & 137.62 & \textbf{119.92} & \textbf{108.06}\\
& Multilingual & \textbf{207.38} & \textbf{158.62} &\textbf{132.49} & 121.23 & 116.52 \\\hline

\multirow{3}{*}{Sw} & LSTM & 355.59 & 293.43 & 226.16 & - & 226.16 \\
& AWD-LSTM & 354.14 & 302.83 & 230.81 & - & 217.14 \\
& Multilingual & \textbf{306.27} & \textbf{250.67} & \textbf{208.87} & - & \textbf{201.20}\\\hline

\multirow{3}{*}{Tl} & LSTM & 214.85 & 160.28 & 131.73 & 122.12 & 124.56 \\
& AWD-LSTM & 291.74 & 177.96 & \textbf{122.63} & \textbf{111.18} & \textbf{102.32}\\
& Multilingual & \textbf{204.29} & \textbf{151.46} & 124.77 & 118.94 & 113.31 \\\hline

\multirow{3}{*}{Tr} & LSTM & 744.41 & 516.49 & 394.46 & 349.26 & 309.58 \\
& AWD-LSTM & 746.09 & 519.40 & 377.47 & \textbf{334.25} & \textbf{303.36}\\
& Multilingual & \textbf{641.14} & \textbf{461.69} & \textbf{371.09} & 340.79 & 308.51 \\\hline
\end{tabular}
\label{tbl:results} 
\end{center}
\vspace{-5mm}
\end{table}

Figure~\ref{fig:result} shows the perplexity curves of the three language models on the test sets of four languages when varying the training set size. The results for five training size thresholds are also reported in Table~\ref{tbl:results}, and the full perplexity scores for each threshold are provided in Appendix~A. 
As can be observed, \texttt{mono-AWD-LSTM} and \texttt{mono-LSTM} perform similarly weak in very low resource settings, while  \texttt{multi-AWD-LSTM} outperforms both by a large margin in all four languages. When training data is sufficiently large, \texttt{mono-AWD-LSTM} achieves the best performance among all models, where \texttt{multi-AWD-LSTM} performs on par with or similar to it. 

Based on our observations, the threshold below which our multilingual model performs better is between 100K to 250K words, depending on the language. The Swahili language in our collection is such a case, as its training data only consists of $\sim$240K words. In this case, \texttt{multi-AWD-LSTM} outperforms all other models even on the full resource setting.

These results show a consistent improvement for our multilingual language models in transferring knowledge across languages when the training data is limited. We attribute this improvement to the parameter sharing at the lower layers, which allows the model to capture language-independent patterns, facilitating better generalization.


\section{Conclusion}
\label{conclusion}
We proposed a novel multilingual language model 
for handling low-resource domains and languages. 
Compared to a state-of-the-art monolingual LM, AWD-LSTM, on four languages, the proposed multilingual LM achieves significant improvements consistently in very low resource scenarios, namely when the size of training data is between 100K to 250K words. The results highlight the benefits of cross-lingual transfer learning for a more effective generalization of LMs on extreme data scarcity scenarios.


\bibliography{main}
\bibliographystyle{acl_natbib}

\appendix
\section{Complete Evaluation Results}
Detail evaluation results on four languages. We refer to our proposed model as Multiling. The other two are monolingual models.  
\begin{table}[h!]
\begin{center}
\caption{Perplexity of LMs in Creole}
\begin{tabular}{c | c c  c }
Training Size & LSTM & AWD-LSTM & Multiling  \\\hline
40K  & 313.64 & 310.41 & \textbf{207.38} \\
60K  & 215.09 & 237.90 & \textbf{185.10} \\
80K  & 195.44 & 201.77 & \textbf{171.63} \\
100K  & 188.96 & 188.18 & \textbf{158.62} \\
120K  & 181.73 & 186.45 & \textbf{150.45} \\
140K  & 158.62 & 150.22 & \textbf{145.16} \\
160K  & 153.50 & 154.11 & \textbf{140.77} \\
180K  & 146.94 & 161.62 & \textbf{136.99} \\
200K  & 144.42 & 137.62 & \textbf{132.49} \\
220K  & 138.62 & 134.34 & \textbf{130.40} \\
240K  & 133.35 & \textbf{120.66} & 126.27 \\
300K  & 127.84 & \textbf{119.92} & 121.23 \\
340K  & 122.88 & \textbf{112.30} & 120.86 \\
400K  & 122.27 & \textbf{108.53} & 117.14 \\
FULL  & 122.27 & \textbf{108.06} & 116.52 \\
\end{tabular}
\label{tbl:res_creole} 
\end{center}
\end{table}
\begin{table}[h!]
\begin{center}
\caption{Perplexity of LMs in Swahili}
\begin{tabular}{c | c c  c }
Training Size & LSTM & AWD-LSTM & Multiling  \\\hline
40K  & 355.59 & 354.14 & \textbf{306.27} \\
60K  & 326.85 & 332.10 & \textbf{277.46} \\
80K  & 314.45 & 314.28 & \textbf{254.52} \\
100K  & 293.43 & 302.83 & \textbf{250.67} \\
120K  & 304.19 & 309.31 & \textbf{237.01} \\
140K  & 249.33 & 259.87 & \textbf{229.69} \\
160K  & 246.57 & 221.59 & \textbf{218.39} \\
180K  & 232.49 & \textbf{211.14} & 213.66 \\
200K  & 226.16 & 230.81 & \textbf{208.87} \\
220K  & 226.16 & 216.12 & \textbf{207.16} \\
FULL  & 226.16 & 217.14 & \textbf{201.20} \\
\end{tabular}
\label{tbl:res_swahili} 
\end{center}
\end{table}

\begin{table}
\begin{center}
\caption{Perplexity of LMs in Tagalog}
\begin{tabular}{c | c c  c }
Training Size & LSTM & AWD-LSTM & Multiling  \\\hline
40K  & 214.85 & 291.74 & \textbf{204.29} \\
60K  & 210.84 & 254.97 & \textbf{173.06} \\
80K  & 251.17 & 204.40 & \textbf{159.58} \\
100K  & 160.28 & 177.96 & \textbf{151.46} \\
120K  & 151.09 & \textbf{145.50} & 145.66 \\
140K  & 145.17 & \textbf{134.92} & 137.97 \\
160K  & 141.40 & \textbf{130.35} & 135.28 \\
180K  & 138.78 & \textbf{129.93} & 130.11 \\
200K  & 131.73 & \textbf{122.63} & 124.77 \\
220K  & 130.38 & 128.44 & \textbf{125.67} \\
240K  & 124.76 & \textbf{118.20} & 122.37 \\
300K  & 122.12 & \textbf{111.18} & 118.94 \\
340K  & 117.39 & \textbf{109.61} & 115.14 \\
400K  & 111.55 & \textbf{103.62} & 110.00 \\
FULL  & 124.56 & \textbf{102.32} & 113.31 \\
\end{tabular}
\label{tbl:res_tagalog} 
\end{center}
\end{table}
\begin{table}
\begin{center}
\caption{Perplexity of LMs in Turkish}
\begin{tabular}{c | c c  c }
Training Size & LSTM & AWD-LSTM & Multiling  \\\hline
40K  & 744.41 & 746.09 & \textbf{641.14} \\
60K  & 635.61 & 733.13 & \textbf{554.16} \\
80K  & 545.93 & 523.87 & \textbf{491.14} \\
100K  & 516.49 & 519.40 & \textbf{461.69} \\
120K  & 477.43 & 466.93 & \textbf{438.43} \\
140K  & 454.03 & 435.06 & \textbf{413.75} \\
160K  & 424.96 & 403.49 & \textbf{393.07} \\
180K  & 415.44 & 387.38 & \textbf{383.95} \\
200K  & 394.46 & 377.47 & \textbf{371.09} \\
220K  & 386.20 & \textbf{362.22} & 363.27 \\
240K  & 363.84 & \textbf{343.20} & 350.35 \\
300K  & 349.26 & \textbf{334.25} & 340.79 \\
340K  & 340.05 & \textbf{316.60} & 329.57 \\
400K  & 315.65 & \textbf{299.17} & 309.18 \\
FULL  & 309.58 & \textbf{303.36} & 308.51 \\
\end{tabular}
\label{tbl:res_turkish} 
\end{center}
\end{table}
\end{document}